# DebateKG – Automatic Policy Debate Case Creation with Semantic Knowledge Graphs

**Allen Roush**
Plailabs
`allen@plailabs.com`

**David Mezzetti**
NeuML
`david.mezzetti@neuml.com`

## Abstract

Recent work within the Argument Mining community has shown the applicability of Natural Language Processing systems for solving problems found within competitive debate. One of the most important tasks within competitive debate is for debaters to create high quality debate cases. We show that effective debate cases can be constructed using constrained shortest path traversals on Argumentative Semantic Knowledge Graphs. We study this potential in the context of a type of American Competitive Debate, called "Policy Debate", which already has a large scale dataset targeting it called "DebateSum". We significantly improve upon DebateSum by introducing 53180 new examples, as well as further useful metadata for every example, to the dataset. We leverage the txtai semantic search and knowledge graph toolchain to produce and contribute 9 semantic knowledge graphs built on this dataset. We create a unique method for evaluating which knowledge graphs are better in the context of producing policy debate cases. A demo which automatically generates debate cases, along with all other code and the Knowledge Graphs, are open-sourced and made available to the public here:
https://huggingface.co/spaces/Hellisotherpeople/DebateKG

## 1 Introduction

### 1.1 Policy Debate

Persuasion has been of interest to humans since we first began communicating with each other. The formal process of using argumentation and rhetoric to convince others to see in one's own way is known as "debate". With varying levels of formality and intensity, these debates happen all around us every day.

More formalized, competitive forms of debate are both highly educational and integral to the formation of a lawful and just society. There is a long and time-honored tradition of academic institutions and news organizations facilitating competitive debate. Many organizations and associations organize debate tournaments according to their differing traditions and rule sets.

Some types of debate are more suited to be assisted with Natural Language Processing systems than others. A popular form of competitive debate done predominantly within United States high schools and universities is called "Policy Debate". Policy Debate maintains one extremely broad and open-ended topic over a whole year, and challenges teams to be ready to either affirm any plan which implements the topic, or to be ready to explain why the opposing teams plan is a bad idea.

Policy Debate is a highly technical form of debate, which puts relatively little emphasis on the aesthetic quality of the speech act, and correspondingly strong emphasis on the quality of the delivered evidence and the delivered argumentation around it. For this reason, Policy Debate rewards teams who can present the maximum amount of evidence possible during their limited speaking time. This leads to a peculiar phenomenon known as "speed reading" or "spreading" which is normalized among most serious competitors. While Policy Debate idiosyncrasies may end up making it less amicable for the general public to watch than other forms, those very same traits make it a uniquely good source of data for NLP systems which generate high quality debate cases.

## 1.2 Policy Debate Cases

Luckily, a large-scale dataset of Policy Debate evidence called DebateSum (Roush and Ballaji., 2020) exists. DebateSum includes all publically available Policy Debate evidence gathered from 2013-2019, which totals to over 180,000 pieces of evidence with corresponding abstractive and extractive summaries alongside rich metadata such as the citation author and word counts.

Beyond its original targeted task of queryable word-level extractive summarization, DebateSum is an excellent dataset for the task of constructing Policy Debate cases. This is because most Policy Debate cases are highly standardized. In almost every Policy Debate round, each debater carefully reads a set of around 3-12 pieces of evidence, starting first with slowly reading the abstractive summary of the evidence (the "argument"), then formulaically reading the evidence citation, and then finally speed reading the extractive summary of the evidence that supports the argument. Moving from each piece of evidence to the next can sometimes be so imperceptible that debaters are instructed to add a slow verbal "next" to their speeches in-between each piece of evidence. Each piece of evidence is likely to be highly related to the previous piece, as they are being chained together to advance the larger narrative of the debate case. This extractive format for debate case construction can be naturally performed by NLP systems which leverage ideas from the Information Retrieval, Graph Analysis, and Distributional Semantics communities.

## 1.3 Semantic Knowledge Graphs

Knowledge Graphs are systems which store information about entities and relates them to each other using (often weighted) edges which show the relationships between each entity. We denote Knowledge Graphs, where each entity consists of documents or sentences, and where weighted edges are constructed between each based on their semantic similarity to each other as "Semantic Knowledge Graphs".

## 1.4 txtai

Computing the semantic similarity between each entity and every other entity is an ideal place to leverage a large scale language model. Approximate Nearest Neighbor (ANN) Systems unlock viable semantic search of these entities, and storing and querying these is a natural place to leverage a database. We are fortunate in that software which does all of these things already exists, and it is called "txtai".

Txtai is a python software package for building AI powered semantic search applications. Txtai features support for a wide variety of backends to power its aforementioned components. Txtai is a natural choice for building Semantic Knowledge Graphs.

## 2 Innovations Introduced

In this work, we introduce several innovations related to automatic Policy Debate case generation.

## 2.1 DebateSum

We significantly improve the existing DebateSum dataset by adding the most recent three additional years of evidence (2020-2022) using the same preprocessing tools as discussed in Roush and Ballaji (2020). This totals to an addition of 53,180 number of documents, bringing the total number of documents within DebateSum to 240,566.

We also add further metadata columns, indicating the source DebateCamp, the broad type of argument, and the topic-year, for all documents within DebateSum. The type of the argument, designated as the "tag", This metadata was extracted from the "openCaselist[1]" project. Figure 1 shows how this metadata was represented on openCaselist.

The additional metadata is particularly useful for more fine-grained information retrieval (e.g. "Give me all evidence about the environment from Gonzaga debate camp in 2013") as well as for leveraging information about the type of debate argument (e.g. "Give me an argument about why individual states should do the plan from the arguments labeled as counterplans").

## 2.2 Contributed Semantic Graphs

We use txtai to build 9 Semantic Knowledge Graphs, which differ based on which column of DebateSum was indexed semantically, and on the language model underlying language model used for similarity calculations. We leave all settings at their defaults during graph construction, which means that networkx is used for the graph backend, huggingface for the language models, faiss for the ANN index, and sqlite for the database. A table of these contributed models is presented in Appendix 1.

---

[1] openCaselist is a continuation of the Open Evidence project and it can be accessed here: https://opencaselist.com/

Figure 1: The added metadata to DebateSum was parsed from tables on openCaselist, which associates each debate document with its camp, its tag (argument types), and its year.

Txtai automatically does topic modeling on each graph using the Louvain (Blondel et al, 2008) community detection algorithm. This data is stored as further information within the graph and unlocks a powerful way to constrain the topics of the generated arguments.

## 2.3 DebateKG

The system that we demonstrate is called "DebateKG". DebateKG is a huggingface "space" webapp which leverages the contributed Semantic Knowledge Graphs to build Policy Debate cases. Users can specify a starting, an ending, and any

1. Warming is real and the product of anthropogenic carbon emissions.
2. Continuing consumption and growth kills the warming
3. Warming prevents drought and famine
4. Warming leads to marine life extinction
5. Extinction – oxygen depletion and food chains
6. Economic growth depletes water resources – recent studies
7. Economic development causes resource depletion
8. Air pollution causes extinction

Figure 2: A Policy Debate Case created with DebateKG. Arguments are shown. The citation, read-aloud extracts, and evidence are omitted for brevity. The first and final argument are the inputs supplied by the user. The highlighted portions show the tokens with the highest similarity to the previous argument, and functions as interpretability.

number of middle arguments. They can also specify any additional constraints, like on the topic, or on the contents of each piece of evidence. DebateKG extracts the evidence closest to the given arguments which meets the given constraints, and then connects these evidence examples together by calculating the constrained weighted shortest path between each evidence example. The portions of each extracted piece of evidence which match the previous portions are highlighted, which functions as a kind of interpretability.

Since there are usually many paths which connect the given pieces of evidence together, there are also many viable debate cases which can be generated. We allow users to generate all possible connected paths (all debate cases), and we enable users to manually display any possible debate case and to interpret the connections between the evidence within them. Besides the automatic case construction functionality, users can also individually query for evidence using txtai's built in semantic SQL language, which helps in the construction of input arguments. Figure 2 shows a sample generated debate case from DebateKG.

## 2.4 Summarization

This work on constructing policy debate cases from semantic knowledge graphs has important implications for summarization tasks. The abstractive and extractive summaries associated with each piece of evidence in DebateSum provide a rich source of training data for abstractive and extractive summarizers. The ability to find constrained shortest paths between evidence examples based on their semantic similarity is analogous to how multi-document summarization aims to find common themes and topics across documents. Additionally, the choice of which text columns to index semantically in the knowledge graphs parallels decisions made in query-focused summarization on what aspects of the text are most relevant to the query. The DebateKG demo itself extracts and concatenates relevant passages into coherent arguments, functioning as an extractive summarization system. The semantic knowledge graph techniques introduced in this work are highly relevant for developing more robust summarization systems. More broadly, the semantic knowledge graph approach introduced here offers representational and algorithmic tools for identifying salient semantics within text collections

## 3 Prior Work

Many others have looked at the relationships between Graph Methods and Argumentation.

The closest prior work to our own comes from IBM Project Debater (Slonim et al., 2021). They created a full debating system which they prominently pitted against champion parliamentary debaters. They defined a custom tailored, "simplified version" of the Parliamentary Debate style. Parliamentary Debate has dramatic differences compared to Policy Debate, namely that the topics are only known to each side 15 minutes ahead of time. As a result, Parliamentary Debate relies far less on evidence, usually only including small snippets as part of a larger speech. In Policy Debate, the vast majority of most of the opening speeches is recitation of extractive summaries of evidence for or against a position. This dramatically simplifies the required system for Policy Debate case generation. Project Debater utilizes many closed source models models, a massive but generalized corpus and requires significantly more compute resources than DebateKG to run.

Finally, Policy Debate is considered to be a more "rigorous style" of debate at its highest level than Parliamentary Debate, which requires dramatically more effort to participate in. An example of this can be found in the 2014-2015 National Parliamentary Tournament of Excellence (NPDA) tournament, the largest American college level parliamentary debate tournament, where the winning team had no prior Parliamentary Debate experience and was otherwise a good but not champion Policy Debate team [2]. Their defeated opponents had been undefeated for the prior 3 years that they competed in the national tournament.

Further work coming from IBM exists about Knowledge Graphs directly being used for Argument Generation (Khatib et al., 2021). Their work explores how to utilize KG encoded knowledge to fine-tune GPT-2 to generate arguments. Our system is extractive in nature, as it creates debate cases by chaining together evidence from DebateSum utilizing graph traversals. Extractive systems are far more appropriate for Policy Debate.

There is fascinating work that applies the idea of Graph Neural Networks for predicting the way that each member of a legislative branch will vote on an input motion (Sawhney et al., 2020). Our work does not try to predict how judges will vote based on any inputs, but instead generates debate cases given input arguments. Their work is in the context of elected officials, whereas ours is in the context high school and collegic competitive debate. There is also work related to trying to understand the arguments made within these legislative Parliamentary Debates (Tamper et al., 2022)

Knowledge Graphs have been utilized for fact checked arguments. ClaimsKG (Tchechmedjiev et al., 2019) is an example, which indexes a wide variety of fact checking websites and annotates them. DebateSum and its contributed KGs do not have fact checking information directly since it is considered the debaters job to convince the judge of the truth of each presented piece of evidence. DebateSum and DebateKG are also significantly larger in size than ClaimsKG and its training corpus.

Work related to automatically evaluating the quality of arguments using Knowledge Graphs exists (Dolz et al., 2022). In their work, they leverage a dataset of debate, the VivesDebate corpus, to identify if an argument is likely to "win". They also recognized the potential for graph traversals to form arguments, or whole debate cases (see figures 2 and 3 from their work). VivesDebate is significantly smaller and less encompassing than DebateSum, and DebateSum does not have information about how successful the arguments within it are.

---

[2] A recording of that final debate round and results can be found here:
https://www.youtube.com/watch?v=l9HJ6Iq6Vas

Other work, which recognizes the potential for paths within knowledge graphs to form arguments, exists (Das et al., 2017). The idea of using "debate dynamics" to present evidence for graph classification has been extensively explored (Hildebrandt et al., 2020). They imagine triple classification and link prediction in graphs as a figurative "debate game" between two reinforcement learning agents who extract "arguments" (paths) which support or oppose a hypothesis. A final binary classifier "judge" votes based on the presented "arguments". They show parallels within Graph Analysis algorithm development to the ideas that we present, but they evaluate this algorithm on non-argumentative datasets. To our knowledge, we are the first work to explore "arguments" (constrained paths) within Knowledge Graphs on an argumentative dataset.

## 4 Details

The DebateKG demo is hosted on huggingface[3]. In this section, we describe the details of DebateKG and its underlying Semantic Knowledge Graphs.

### 4.1 Underlying Language Models

Txtai supports several language modeling backends, the most modern of which is sentence transformers (Reimers and Gurevych., 2019). Besides having many pre-trained language models which are designed for Semantic Textual Similarity or for Sentence Modeling, any Transformer model can be transformed into a "sentence transformer" model with nothing more than a pooling layer added.

We choose three language models for building the Knowledge Graphs. The first is the recommended model from the sentence transformers documentation [4], "all-mpnet-base-v2". We are also curious about the potential usefulness of language models which are fine-tuned in a domain similar to DebateSum, such as the legal domain. We choose "legal-bert-base-uncased" (Chalkidis et al., 2020) for this reason, as it is trained on a diverse legal corpus. Finally, we are curious about language models which can model long sequences. We choose "allenai/longformer-base-4096" (Beltagy et al., 2020) due to its potential to model sequences up to 4096 tokens long directly.

### 4.2 Importance of Granularity

For each piece of evidence in DebateSum, there is an associated abstractive summary and biased extractive summary. Since at the time of writing, txtai and DebateKG can only semantically index one text column at a time, the choice of which column and at what granularity is highly important. There are merits and drawbacks to each approach. For this reason, we construct Graphs which index two of these columns (denoted with the prefixes "DebateKG-ext", and "DebateKG-abs"). We also construct graphs which index each individual sentence of the full document (denoted as "DebateKG-sent"). These graphs are significantly larger, but are potentially far more potent since the sentence transformers recommended models are designed for the sentence granularity and because the other two models are average pooled and subsequently long sequences dilute their embeddings.

### 4.3 Importance of Settings

DebateKG computes the semantic similarity between each entity, and connects the entities whose similarity is greater than a user-defined threshold. We use the default threshold of 0.10, and each entity has a limit of no more than 100 edges. Changes in these settings, such as lowering the threshold and increasing the entity limit, will result in more highly connected and correspondingly larger graphs.

### 4.4 Policy Debate Case Construction

The shortest paths, which minimizes the semantic distance between each input argument, are also

---

[3] The link to that demo is here: https://huggingface.co/spaces/Hellisotherpeople/DebateKG

[4] An analysis of the pretrained models can be found here: https://www.sbert.net/docs/pretrained_models.html

Policy Debate Arguments[5]. One or more of these Arguments can be concatenated to form Policy Debate Cases. The ideal Policy Debate Argument uses the minimum amount of spoken words. This enables competitors to make more arguments, and to make broader and stronger cases.

Beyond a naïve shortest path calculation on the whole graph, we can control how Debate Case are constructed by choosing to run these calculations on subgraphs. These subgraphs include only entities which fulfil a particular constraint – enabling things like arguments where all of the evidence stays on a particular topic, or which always includes a keyword, or even where the evidence isn't longer than a certain number of words.

Related to the idea of minimizing the number of words spoken out loud within each debate case, we can also modify the scoring function used within the shortest path calculations to account for and try to minimize the length of the evidences extracts. This has the advantage over selecting subgraphs of allowing for inclusion of long documents within the argument if they are actually the most appropriate.

### 4.5 Value of Knowledge Graphs

While an exhaustive analysis of these Knowledge Graphs is beyond the scope of this paper, it is important to recognize that techniques and algorithms from the Graph Analysis literature can be particularly illuminating. Centrality algorithms, like Pagerank (Page et al., 1998), will find evidence which is highly applicable to many arguments. Community detection, also known as clustering – finds evidence which is highly related to each other. A treasure trove of insights into DebateSum are unlocked for those willing to explore the Semantic Knowledge Graphs.

### 4.6 Connection to the Legal Domain

Policy debate focuses extensively on legal topics, as debaters must argue matters of policy frequently connected to the law. As such, the DebateSum dataset provides a trove of legal text spanning diverse issues. The constrained shortest path approach introduced mirrors legal research in finding related precedents and documents to build a broader argument. The graph structure encodes useful semantics for legal information retrieval, while segmentation by sentence allows fine-grained modeling of argument components within judicial opinions.

---

[5] And in fact, any path on this graph can be an Argument

| Model | Average Words in Case |
|---|---|
| Mpnet-DebateKG-abs | 406 |
| Mpnet-DebateKG-ext | 305 |
| Mpnet-DebateKG-sent | 760 |
| legalbert-DebateKG-abs | 502 |
| legalbert-DebateKG-ext | **230** |
| legalbert-DebateKG-sent | 709 |
| longformer-DebateKG-abs | 500 |
| longformer-DebateKG-ext | 457 |
| longformer-DebateKG-sent | 301 |

Table 1: Results of experiment on sample 10 arguments

Further, the legal-domain language model explores domain adaptation of BERT models, an important avenue in legal NLP. The interpretable paths can explain how fragments are related, promoting model transparency important for legal applications. Graph-based representation and algorithms for semantic similarity offer techniques to push forward core legal NLP tasks. With rich legal data and domain-specific modeling, this work makes both methodological and data contributions highly relevant for legal NLP. The techniques could further applications like legal search, summarization of contracts or cases, and argument mining over caselaw.

## 5 Evaluation

DebateSum does not include any data indicating if an argument is "strong", or if it is likely to win or not. It also does not have similarity labels between each example or even between pairs of samples. This means that it is challenging to compare the argumentation quality of each graph. Fortunately, it is simple to look at the lengths of the spoken aloud extracts. Since Policy Debaters are trying to minimize the time spent on each argument, they will prefer Graphs that extract evidence chains with shorter extracts.

Thus, we evaluate each graph based on how long the created Debate Cases extracts are. We choose 10 input argument pairs (a table of which is included within the github repo) and rank each graph based on the average length of the read aloud extracts from the generated debate cases across all 10 of these argument pairs. Table 1 shows the results of this experiment.

Due to the unique and small-scale nature of our evaluation, we hope that future work can find more effective ways to evaluate Semantic Knowledge Graphs in an argumentative context.

## 6 Conclusion

In this paper, we significantly expanded and improved an existing large scale argument mining dataset called "DebateSum". We created 9 Semantic Knowledge Graphs using the "txtai" Semantic AI toolkit. We showed how constrained shortest path traversals on these graphs can be used to create Policy Debate Cases. We created a System Demonstration of this called DebateKG which is a "space" webapp hosted on huggingface. We discuss implementation details of this system. We propose a way for Policy Debaters to decide which graph is better for their needs, and evaluate our systems using this technique. We open source all data, code, and graphs.

## Limitations

The largest of the contributed Semantic Graphs, denoted "DebateKG-sent", can require as much as 100gb of free-space on disk when uncompressed (which is required to leverage them). All training and creation of these graphs was performed on a personal computer with an RTX 3080ti GPU, an I7 8700K CPU, and 32gigs of ram.

American Policy Debate, is almost always performed in English, and it is unlikely that suitable training data targeting it outside of English will be created in the near future.

DebateSum is crowd sourced from high school and college Policy Debate camp attendees. The evidence found within DebateSum, as well as the additions included within this paper, may have some annotation and/or parsing errors. This is because while the general layout of evidence is agreed upon by all, there is much variance in the formatting.

## Ethics Statement

Philosophy, Law, Politics, Economics, and other Social Sciences are particularly well represented within DebateSum due to its nature as an argumentative dataset. The Policy Debate community has strong norms and supervision related to the included content which make the risk of hurtful or harmful content being included to be low. Still, the possibility of problematic content being included cannot be fully eliminated.

DebateKG is an extractive system. While extractive systems have far lower abuse potential compared to generative systems, the risk of abuse is also not totally eliminated. A "dialectic", according to the ancient philosopher Plato, is a dialogue held between two or more people for the purposes of finding truth. By contrast, a "debate", as far as competitors are concerned, is nothing more than a game of rhetorical persuasion played with real life evidence and situations. While most evidence within DebateSum is fully cited and is generally high quality, the way that that the evidence is summarized is biased towards the targeted argument that the competitor was trying to craft.

We also point out that DebateSum is not necessarily factual or "truthful". While the evidence within it should have almost no direct "lies", "fabrications" or "fake-news", the evidence can still be misleading or without important context.

## References

Allen Roush and Arvind Balaji. 2020. DebateSum: A large-scale argument mining and summarization dataset. In *Proceedings of the 7th Workshop on Argument Mining*, pages 1–7, Online. Association for Computational Linguistics.

Vincent D. Blondel, Jean-Loup Guillaume, Renaud Lambiotte, Etienne Lefebvre. 2008. Fast unfolding of communities in large networks. In Journal of Statistical Mechanics, pages 1-8

Slonim, N., Bilu, Y., Alzate, C. et al. 2021. An autonomous debating system. In *Nature*, pages 379–384.

Khalid Al Khatib, Lukas Trautner, Henning Wachsmuth, Yufang Hou, and Benno Stein. 2021.

Employing Argumentation Knowledge Graphs for Neural Argument Generation. In *Proceedings of the 59th Annual Meeting of the Association for Computational Linguistics and the 11th International Joint Conference on Natural Language Processing (Volume 1: Long Papers)*, pages 4744–4754, Online. Association for Computational Linguistics.

Ramit Sawhney, Arnav Wadhwa, Shivam Agarwal, and Rajiv Ratn Shah. 2020. GPolS: A Contextual Graph-Based Language Model for Analyzing Parliamentary Debates and Political Cohesion. In *Proceedings of the 28th International Conference on Computational Linguistics*, pages 4847–4859, Barcelona, Spain (Online). International Committee on Computational Linguistics.

Minna Tamper, Rafael Leal, Laura Sinikallio, Petri Leskinen, Jouni Tuominen, and Eero Hyvonen. 2022. Extracting Knowledge from Parliamentary Debates for Studying Political Culture and Language. Online

Tchechmedjiev, A., Fafalios, P., Boland, K., Gasquet, M., Zloch, M., Zapilko, B., Dietze, S., Todorov, K., ClaimsKG: A Live Knowledge Graph of Fact-Checked Claims. 2019. In *18th International Semantic Web Conference (ISWC19)*, Auckland, New Zealand,

Ruiz-Dolz, R., Heras, S., & García-Fornes, A. 2022. Automatic Debate Evaluation with Argumentation Semantics and Natural Language Argument Graph Networks. *arXiv preprint arXiv:2203.14647*.

Rajarshi Das, Arvind Neelakantan, David Belanger, and Andrew McCallum. 2017. Chains of Reasoning over Entities, Relations, and Text using Recurrent Neural Networks. In *Proceedings of the 15th Conference of the European Chapter of the Association for Computational Linguistics: Volume 1, Long Papers*, pages 132–141, Valencia, Spain. Association for Computational Linguistics.

Hildebrandt, Marcel & Serna, Jorge & Ma, Yunpu & Ringsquandl, Martin & Joblin, Mitchell & Tresp, Volker. 2020. Debate Dynamics for Human-comprehensible Fact-checking on Knowledge Graphs. *In AAAI 2019 Fall Symposium Series*

Nils Reimers and Iryna Gurevych. 2019. Sentence-BERT: Sentence Embeddings using Siamese BERT-Networks. In *Proceedings of the 2019 Conference on Empirical Methods in Natural Language Processing and the 9th International Joint Conference on Natural Language Processing (EMNLP-IJCNLP)*, pages 3982–3992, Hong Kong, China. Association for Computational Linguistics.

Ilias Chalkidis, Manos Fergadiotis, Prodromos Malakasiotis, Nikolaos Aletras, and Ion Androutsopoulos. 2020. LEGAL-BERT: The Muppets straight out of Law School. In *Findings of the Association for Computational Linguistics: EMNLP 2020*, pages 2898–2904, Online. Association for Computational Linguistics.

Iz Beltagy, Matthew E. Peters, Arman Cohan. 2020. Longformer: The Long-Document Transformer, *arXiv*.

Page, Lawrence and Brin, Sergey and Motwani, Rajeev and Winograd, Terry. 1999. The PageRank Citation Ranking: Bringing Order to the Web. Technical Report. *Stanford InfoLab*.

## A  Appendix 1: Table of Contributed Models

| Model Name | Number of Vertices | Number of Edges | Average Degree |
|---|---|---|---|
| Mpnet-abs | 240566 | 1876918 | 7.80 |
| Mpnet-ext | 240566 | 2133792 | 8.86 |
| Mpnet-sent | 2546059 | 68305930 | 19.3 |
| Legalbert-abs | 240566 | 3006572 | 11.16 |
| Legalbert-ext | 240566 | 2685362 | 12.49 |
| Legalbert-sent | 2546059 | 48352931 | 21.5 |
| Longformer-abs | 240566 | 3685467 | 6.56 |
| Longformer-ext | 240566 | 5507938 | 8.89 |
| Longformer-sent | 2546059 | 59743621 | 22.4 |